\documentclass[10pt,twocolumn,letterpaper]{article}

\usepackage{iccv}
\usepackage{times}
\usepackage{caption}
\usepackage{subcaption}
\usepackage{epsfig}
\usepackage{graphicx}
\usepackage{amsmath}
\usepackage{amssymb}
\usepackage{array}
\usepackage{colortbl}
\usepackage{booktabs}
\usepackage{bbm}
\usepackage{multirow}
\usepackage{multicol}
\usepackage{makecell}
\usepackage{xcolor}
\usepackage{xspace}
\usepackage{float}
\usepackage{color}
\usepackage{algorithm}
\usepackage{listings}

\newcommand{\myparagraph}[1]{{\vspace{.5em} \noindent \bf #1}}

\usepackage[pagebackref=true,breaklinks=true,letterpaper=true,colorlinks,bookmarks=false]{hyperref}

\iccvfinalcopy 

\def\iccvPaperID{0000}

\begin{document}

\title{TransTrack: Multiple Object Tracking with Transformer}

\author
{
Peize Sun$^{1}$, 
~~
Jinkun Cao$^{2}$,
~~
Yi Jiang$^{3}$,
~~
Rufeng Zhang$^{4}$,
~~
Enze Xie$^{1}$, \\
~~
Zehuan Yuan$^{3}$, 
~~~
Changhu Wang$^{3}$, 
~~~
Ping Luo$^{1}$
\\[0.2cm]
${^1}$The University of Hong Kong ~~~
${^2}$Carnegie Mellon University\\ 
${^3}$ByteDance AI Lab ~~~
${^4}$Tongji University ~~~
}

\maketitle

\begin{abstract}
In this work, we propose \textbf{TransTrack}, a simple but efficient scheme to solve the multiple object tracking problems. TransTrack leverages the transformer architecture, which is an attention-based query-key mechanism. 
It applies object features from the previous frame as a query of the current frame and introduces a set of learned object queries to enable detecting new-coming objects.
It builds up a novel joint-detection-and-tracking paradigm by accomplishing object detection and object association in a single shot, simplifying complicated multi-step settings in tracking-by-detection methods. On MOT17 and MOT20 benchmark, TransTrack achieves 74.5\% and 64.5\% MOTA, respectively, competitive to the state-of-the-art methods. We expect TransTrack to provide a novel perspective for multiple object tracking.
The code is available at: {\footnotesize\url{https://github.com/PeizeSun/TransTrack}.}

\vspace{-1mm}
\end{abstract}

\section{Introduction}
\begin{figure}[t]
\begin{center}
\begin{subfigure}{0.49\textwidth}
    \centering
    \includegraphics[width=0.85\textwidth]{ 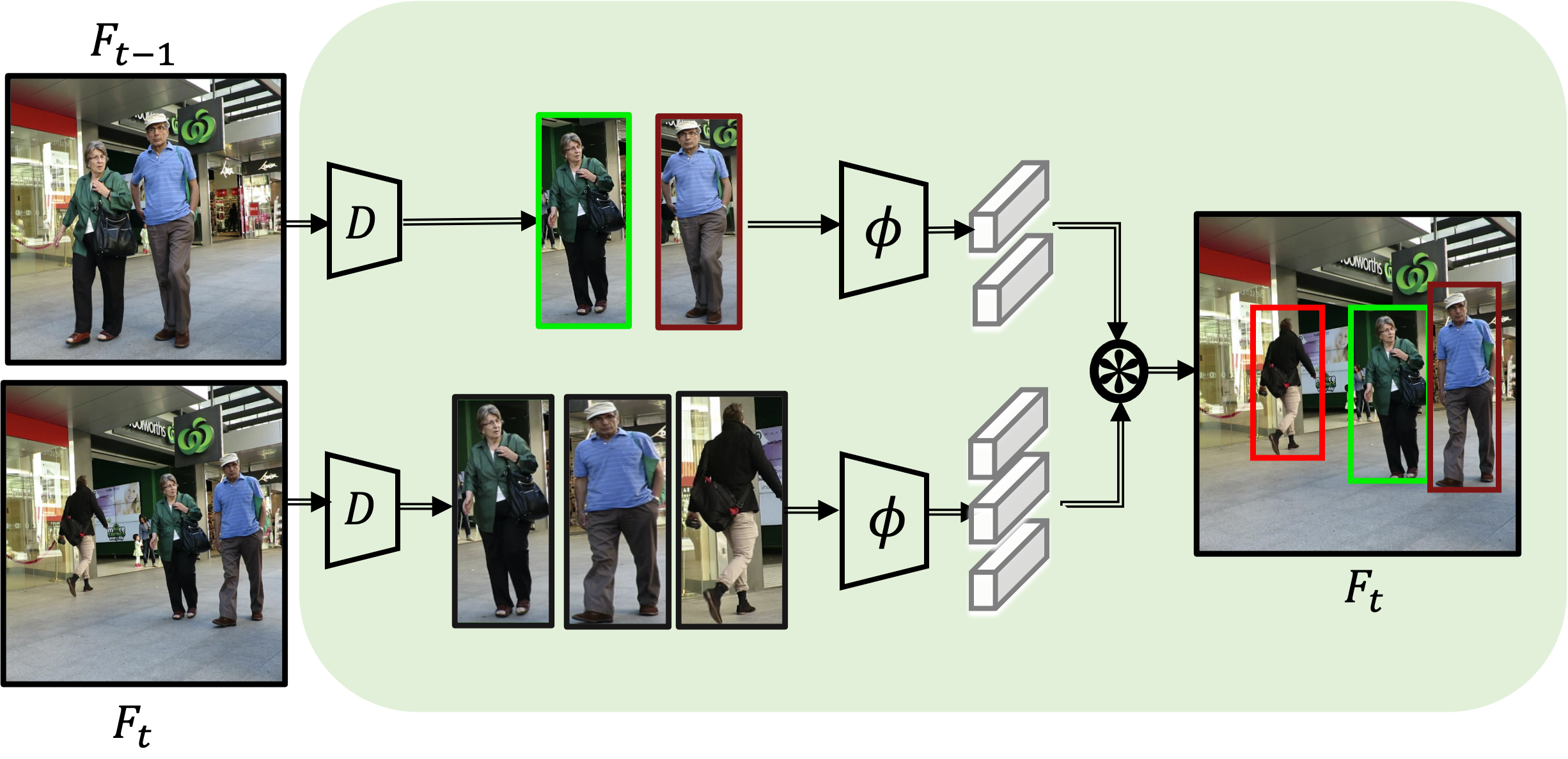}
    \vspace{-1mm}
    \caption{Complex tracking-by-detection MOT pipeline.}
    \label{fig:1a}	
\end{subfigure}
\vspace{2mm}
    
\begin{subfigure}{0.49\textwidth}
    \centering
    \includegraphics[width=0.85\textwidth]{ 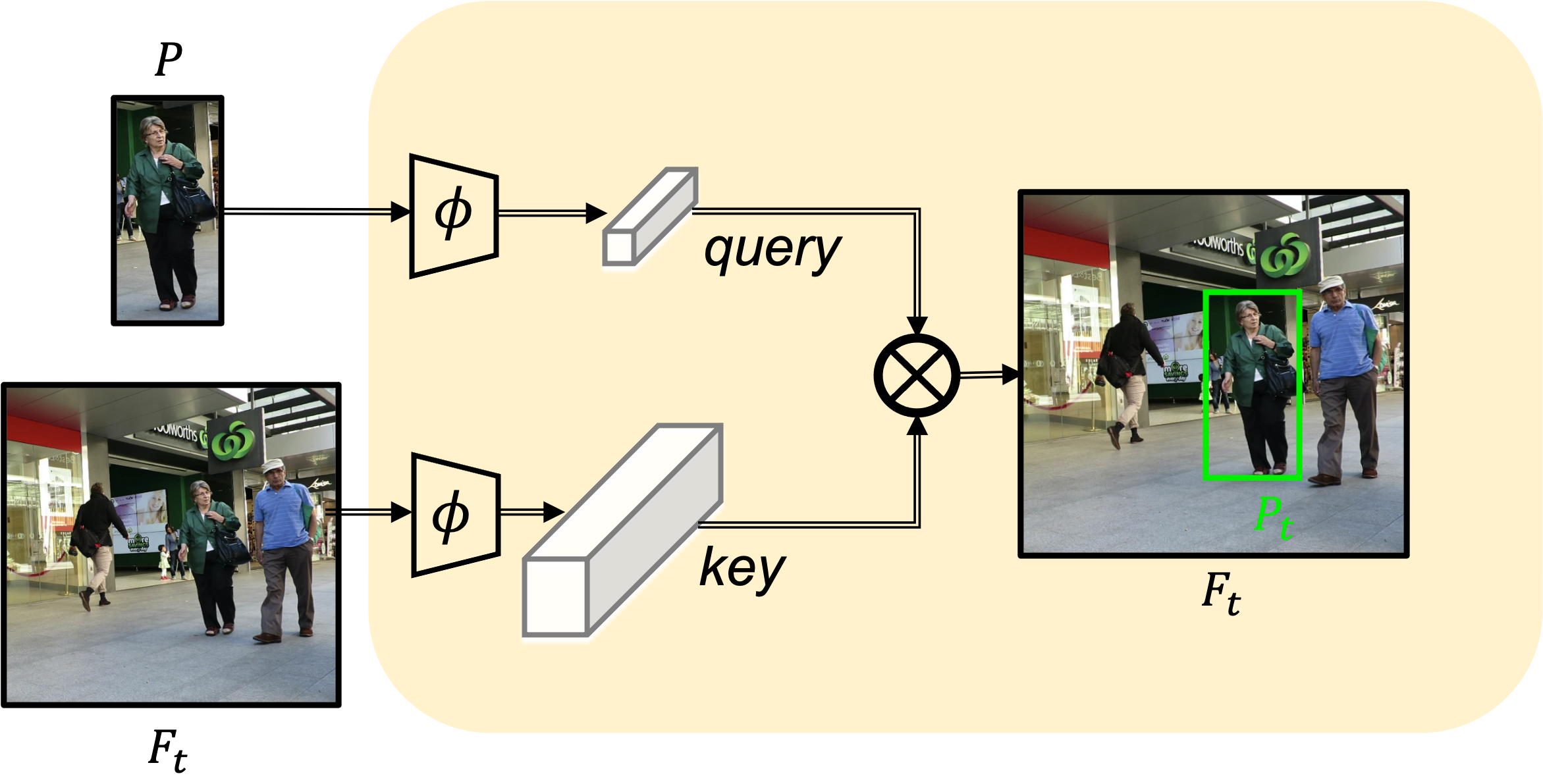}
    \vspace{-2mm}
    \caption{Simple query-key SOT pipeline.}
    \label{fig:1b}	
\end{subfigure}
\vspace{2mm}

\begin{subfigure}{0.49\textwidth}
    \centering
    \includegraphics[width=0.85\textwidth]{ 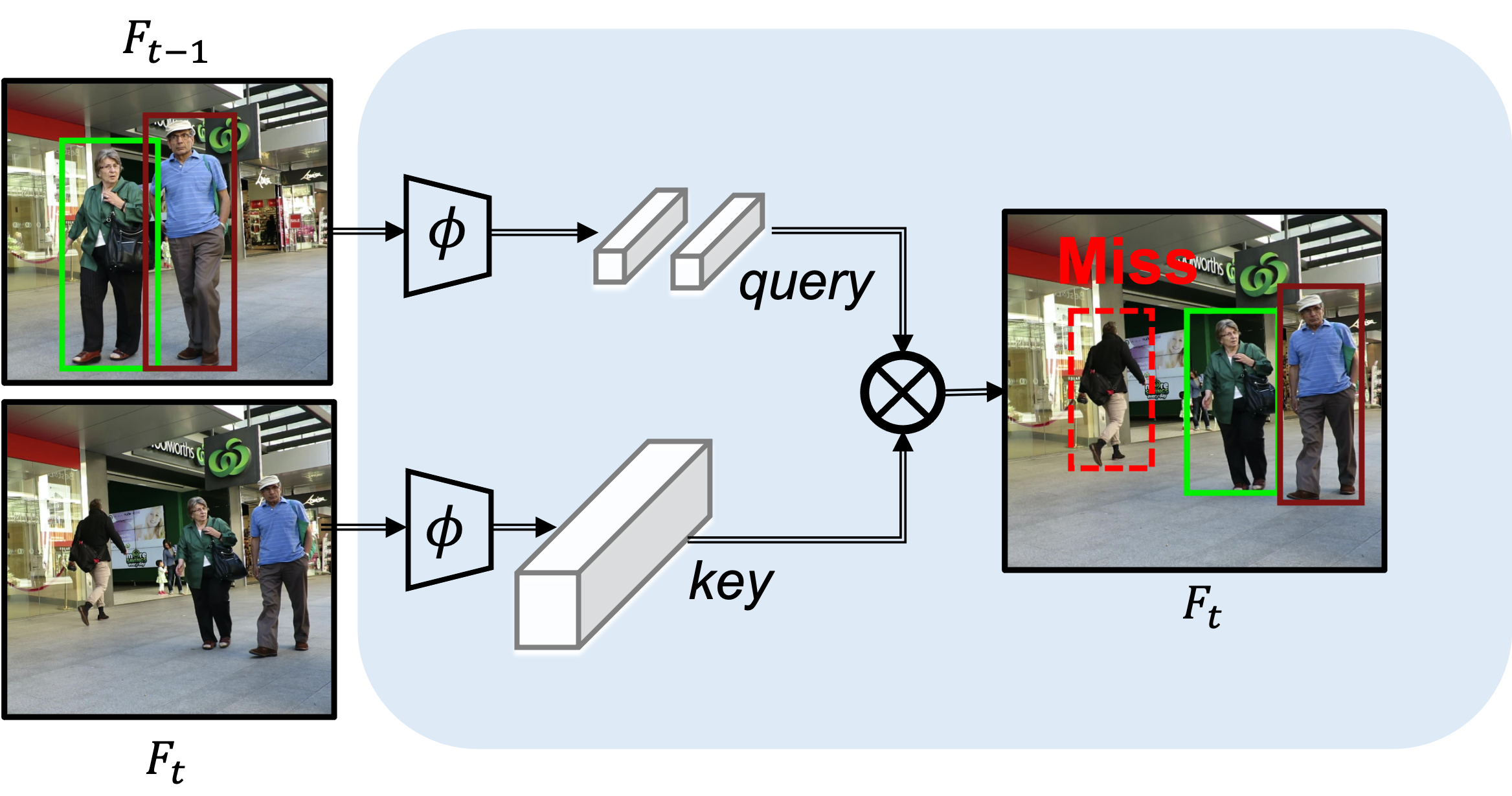}
    \vspace{-2mm}
    \caption{Query-key pipeline has great potential to setup a simple MOT method. However, it will \textbf{miss} new-coming objects.}
    \label{fig:1c}	
\end{subfigure}

\caption{\textbf{Motivation of TransTrack}. The dominant MOT method is the complex multi-step tracking-by-detection pipeline. Directly migrating the query-key mechanism from SOT to MOT will cause severe missing of new-coming objects. TransTrack is aimed to take advantage of query-key mechanism and to detect new-coming objects. The pipeline is shown in Figure~\ref{fig:pipeline}.
}
\label{fig:motivation}
\end{center}
\vspace{-1cm}
\end{figure}

\begin{figure*}[htbp]
\begin{center}
\includegraphics[width=0.95\textwidth]{ 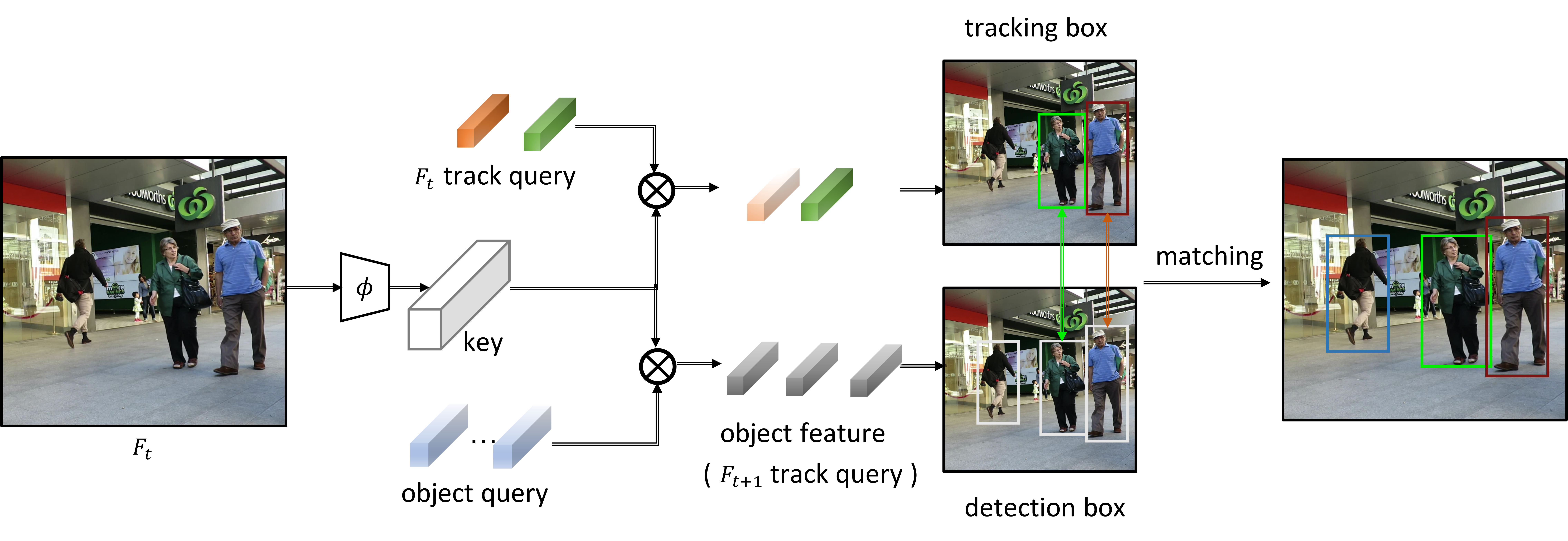}
\caption{\textbf{Pipeline of TransTrack}. Both object feature query from the previous frame and learned object queries are taken as input. The image feature maps are a shared key. The learned object query detects objects in the current frame. The track query from the previous frame associates objects of the current frame with the previous ones. This process is performed sequentially over all adjacent frames and finally completes the multiple object tracking tasks.} 
\label{fig:pipeline}
\end{center}
\vspace{-7mm}
\end{figure*}

Visual object tracking is a vital problem in many practical applications, such as visual surveillance, public security, video analysis, and human-computer interaction. According to the number of objects to track, the task of object tracking is divided into \textbf{Single Object Tracking (SOT)} and \textbf{Multiple Object Tracking (MOT)}. In recent years, the emerging of deep siamese networks~\cite{bertinetto2016fullyconvolutional, tao2016siamese, siamRPN, li2018siamrpn} have made great progress in solving SOT tasks. However, the existing MOT methods are still suffering from the model complexity and computational cost due to the multi-stage pipeline~\cite{yu2016poi, Tang_2017_CVPR, wang2019towards} as shown in Figure~\ref{fig:1a}. 

A critical dilemma in many existing MOT solutions is when object detection and re-identification are performed separately, they can not benefit each other. To tackle the problem in MOT, a joint-detection-and-tracking framework is needed to share knowledge between detection and object association. By reviewing SOT solutions, we emphasize that \textbf{Query-Key} mechanism is promising in this direction. In existing works, the object target is the query and the image regions are the keys as shown in Figure~\ref{fig:1b}. For the same object, its feature in different frames is highly similar, which enables the query-key mechanism to output ordered object sets. This inspiration should also be beneficial to the MOT task.

However, merely transferring the vanilla query-key mechanism from SOT into the MOT task leads to poor performance, significantly causing much more false negatives. It is because when an new object comes into birth, there is no corresponding features for it. This defect causes severe object missing, as shown in Figure~\ref{fig:1c}. So what is a suitable query-key mechanism for MOT remains a critical question. A desirable solution should be able to well capture new-coming objects and propagate previously detected objects to the following frames at the same time.

In this paper, we make efforts in this direction by building an MOT framework based on transformer~\cite{vaswani2017attention}, which is an attention-based query-key mechanism. We term it as \textbf{TransTrack}. It leverages set prediction for detection~\cite{DETR} and the knowledge passed from the previous frame to gain reliable object association at the same time. There are two sets of keys (following previous works~\cite{DETR}, they are confusingly termed as ``object query'' in transformer). One  set contains the object queries learned as in existing transformer-based detector~\cite{DETR} and the other contains those generated from the features of objects on the previous frame, which are also termed as ``track query'' for clarification. The first set of queries provides a sense of new-coming objects and the track queries provide consistent object information to maintain tracklets. Two sets of bounding boxes are predicted respectively and TransTrack uses simple IoU matching to generate the final ordered object set from them.

In TransTrack, the two sets of boxes can be output from a uniform decoder architecture with only different queries as input. Our model even removes the traditional NMS stage in detection. Therefore, our method is simple and straightforward where all components of the model can be trained at the same time. We evaluate TransTrack on the two real-world benchmarks MOT17 and MOT20~\cite{mot16,mot20}. It achieves 74.5 and 64.5 MOTA on the test set of MOT17 and MOT20 respectively. To the best of our knowledge, we are the first to introduce the transformer in the MOT task. As it has achieved comparable performance with state-of-the-art models, we hope it could provide a new perspective and efficient baseline for multi-object tracking tasks.

\section{Related Work}
In this section, we first review previous transformer applications in vision tasks. Then we introduce the two main MOT paradigms, namely tracking-by-detection and joint-detection-and-tracking methods.

\myparagraph{Transformer in vision tasks.} Recently, there is a popularity of using transformer architecture~\cite{vaswani2017attention} in vision tasks, where it has been proven powerful and inspiring. As a special query-key mechanism, the transformer heavily relies on the attention mechanism to process extracted deep features. It first shows great efficiency in natural language processing~\cite{vaswani2017attention} and later migrated to visual perception tasks~\cite{DETR} achieving remarkable success. Transformer appeals to the vision community with elegant structure and good performance. It has shown great potential in detection~\cite{DETR, deformdetr}, segmentation~\cite{zheng2020rethinking}, 3D data processing~\cite{zhao2020point} and even backbone construction~\cite{dosovitskiy2020image}. Lately, the good efforts of using a transformer in processing sequential visual data also make remarkable shots in video segmentation~\cite{wang2020endtoend}. With the natural strength of passing features along the temporal dimension, the transformer shows the ability to contribute to diverse temporal-spatial processing tasks on visual data and even replaces the role of traditional RNN models ~\cite{hochreiter1997long}. However, to the best of our knowledge, there are still no published transformer-based solutions for object tracking while it is intuitive to leverage its demonstrated good capacity in visual perception and temporal processing there. Hence, in this paper, we follow the insight to propose a transformer-based model for MOT. It shows convincingly high performance on the popular MOT benchmark.

\myparagraph{Tracking-by-detection.} State-of-the-art multiple object trackers are mostly dominated by the tracking-by-detection paradigm. It ﬁrstly uses the object detectors~\cite{lin2018focal, ren2016faster, FPN} to localize all objects of interest, then associates these detected objects according to their Re-ID features and/or other information, e.g., Intersection over Unions (IoU) between each other. SORT~\cite{Bewley2016_sort} tracks bounding boxes using the Kalman Filter~\cite{welch1995introduction} and associates to the current frame by the Hungarian algorithm~\cite{kuhn1955hungarian}. DeepSORT~\cite{wojke2017simple} replaces the association cost in SORT with the appearance features from deep convolutional networks. POI~\cite{yu2016poi} achieves state-of-the-art tracking performance based on the high-performance detection and deep learning-based appearance feature. Lifted-Multicut~\cite{Tang_2017_CVPR} combines the deep representations and body pose feature obtained by the pose estimation model. STRN~\cite{xu2019spatialtemporal} presents a similarity learning framework between tracks and objects, which encodes various Spatial-Temporal relations. Tracking-by-detection pipeline achieves leading performance, but its model complexity and computational cost are not satisfying. 

\myparagraph{Joint-detection-and-tracking.}  The joint-detection-and-tracking pipeline aims to achieve detection and tracking simultaneously in a single stage. D\&T~\cite{feichtenhofer2018detect} proposes a multi-task architecture for frame-based object detection and across-frame track regression. Integrated-Detection~\cite{zhang2018integrated} boosts the detection performance by combining the detection bounding boxes in the current frame and tracks in previous frames. More recently, Tracktor~\cite{bergmann2019tracking} directly uses the previous frame tracking boxes as region proposals and then applies the bounding box regression to provide tracking boxes on the current step, thus eliminating the box association procedure. JDE~\cite{wang2019towards} and FairMOT~\cite{fairmot} learn the object detection task and appearance embedding task from a shared backbone. CenterTrack~\cite{zhou2020tracking} localizes objects by tracking-conditioned detection and predicts their offsets to the previous frame. ChainedTracker~\cite{peng2020chained} chains paired bounding boxes estimated from overlapping nodes, in which each node covers two adjacent frames.
Our proposed TransTrack falls into the joint-detection-and-tracking category. Previous works adopt anchor-based~\cite{ren2016faster} or point-based~\cite{zhou2019objects} detection framework. Instead, we build the pipeline based on a query-key mechanism and the tracked object feature is used as the query. 

\begin{figure}[!t]
\begin{center}
\includegraphics[width=0.48\textwidth]{ 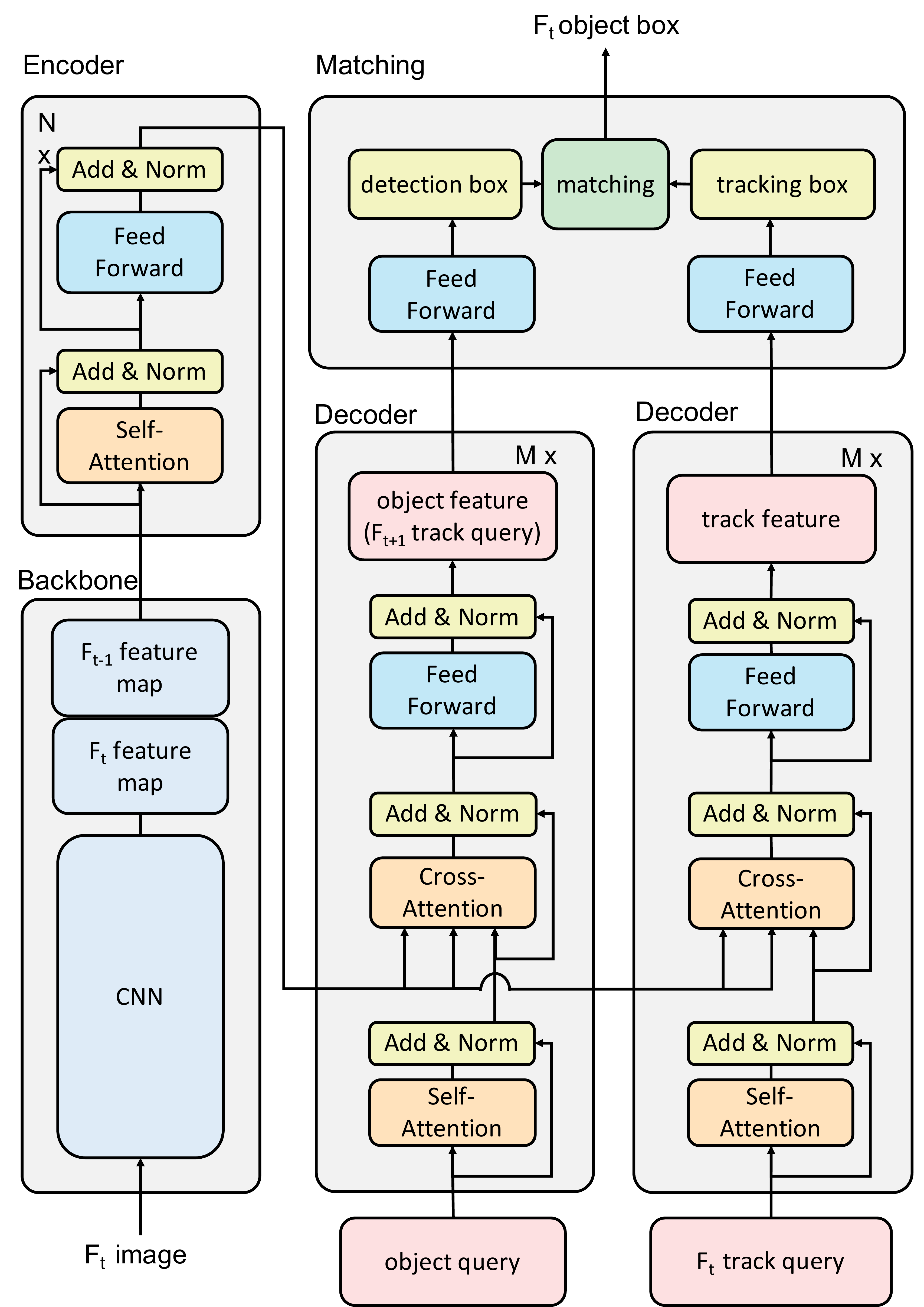}
\vspace{-3mm}
\caption{\textbf{The architecture details of TransTrack}. First, the current frame image is input to CNN backbone to extract feature map. Then, both the current frame feature map and the previous one are fed into encoder to generate composite feature. Next, learned object query is decoded into detection boxes and object feature of the previous frame is decoded into tracking boxes. Finally, IoU matching is employed to associate detection boxes to tracking boxes.}
\label{fig:detail}
\end{center}
\vspace{-3mm}
\end{figure}

\section{TransTrack}
In MOT task, the desirable output is a \textbf{complete} and \textbf{correctly ordered} set of objects on each frame in a video. To these two ends, TransTrack uses queries from two sources to gain adaptive cues. On the one hand, similar to usual transformer-based detectors~\cite{DETR, deformdetr}, TransTrack takes an object query as input to provide common object detection results. On the other hand, TransTrack leverages features from previously detected objects to form another ``track query'' to discover associated objects on the following frames. Under this scheme, TransTrack generates in parallel two sets of bounding boxes, termed as ``detection boxes'' and ``tracking boxes''. Last, TransTrack uses the Hungarian algorithm, where the cost is IoU area among boxes, to achieve the final ordered box set from the two bounding box sets. The pipeline is illustrated in Figure~\ref{fig:detail}.

\vspace{2mm}
\subsection{Pipeline}
In this section, we introduce the encoder-decoder architecture of TransTrack for object detection and object propagation. Given the detection boxes and tracking boxes from two decoders, box IoU matching is used to obtain the final tracking result. We also introduce the training and inference process of TransTrack.

\vspace{1mm}
\noindent\textbf{Architecture.} 
TransTrack is based on transformer, an encoder-decoder framework. 
It replies on stacked multi-head attention layers and feed-forward networks. Multi-head attention is called self-attention if the input query and the input key are the same, otherwise, cross-attention. 
In transformer architecture, The encoder generates keys and the decoder takes as input task-specific queries. 
The architecture overview is shown in Figure~\ref{fig:detail}.

The encoder of TransTrack takes the composed feature maps of two consecutive frames as input. To avoid duplicated computation, the extracted features of the current frame are temporarily saved and then re-used for the next frame. 
Two parallel decoders are employed in TransTrack. Feature maps generated from the encoder are used as common keys by the two decoders. The two decoders are designed to perform object detection and object propagation, respectively. 
Specifically, a decoder takes learned object query as input and predicts \textit{detection boxes}. The other decoder takes the object feature from previous frames, namely ``track query'', as input and predicts the locations of the corresponding objects on the current frame, whose bounding boxes are termed as \textit{tracking boxes}.

\vspace{1mm}
\noindent\textbf{Object Detection.} 
Following DETR~\cite{DETR}, TransTrack leverages learned object query for object detection. The object query is a set of learnable parameters, trained together with all other parameters in the network. During detection, the key is the global feature maps generated from the input image and the object query looks up objects of interest in the image and outputs the final detection predictions, termed as  ``detection boxes''. This stage is performed by the left-hand decoder block in Figure~\ref{fig:detail}.

\vspace{1mm}
\noindent\textbf{Object Propagation.}
Given detected objects in the previous frame, TransTrack propagates these objects by passing their features to the next frame as the track query. The stage is performed by the right-hand decoder block in Figure~\ref{fig:detail}. The decoder has the same architecture as the left-hand one but takes queries from different sources. This inherited object feature conveys the appearance and location information of previously seen objects, so this decoder could well locate the position of the corresponding object on the current frame and output ``tracking boxes''.

\vspace{2mm}
\noindent\textbf{Box Association.}
Provided the detection boxes and tracking boxes, TransTrack uses the box IoU matching method to get the final tracking result, as shown in Figure~\ref{fig:detail}. 
Applying the Kuhn-Munkres (KM) algorithm~\cite{kuhn1955hungarian} to IoU similarity of detection boxes and tracking boxes, detection boxes are matched to tracking boxes. Those unmatched detection boxes are kept to create new tracklets.

\subsection{Training}
\noindent\textbf{Training Data.}
We build training dataset from two sources. As usual, the training data of could be two consecutive frames or two randomly selected frames from a real video clip.
Furthermore, training data could also be the static image~\cite{zhou2020tracking}, where the adjacent frame is simulated by randomly scaling and translating the static image.

\vspace{2mm}
\noindent\textbf{Training Loss.}
In TransTrack, tracking boxes and detection boxes are the predictions of object boxes in the same image. It allows us to simultaneously train two decoders by the same training loss.

TransTrack applies set prediction loss to supervise detection boxes and tracking boxes of classification and box coordinates. Set-based loss produces an optimal bipartite matching between predictions and ground truth objects. Following
~\cite{DETR, deformdetr,sun2020sparse,sun2020onenet,wang2020end}, the matching cost is defined as
\begin{equation}
    \label{total_loss}
    \mathcal{L} = \lambda_{cls}  \cdot  \mathcal{L}_{\mathit{cls}} + \lambda_{L1} \cdot \mathcal{L}_{\mathit{L1}} +
    \lambda_{giou} \cdot \mathcal{L}_{\mathit{giou}}
\end{equation}
where
$\mathcal{L}_{\mathit{cls}}$ is focal loss~\cite{lin2018focal} of predicted classifications and ground truth category labels, $\mathcal{L}_{\mathit{L1}}$ and $\mathcal{L}_{\mathit{giou}}$ are L1 loss and generalized IoU loss~\cite{GIoU} between normalized center coordinates and height and width of predicted boxes and ground truth box, respectively. 
$\lambda_{cls}$, $\lambda_{L1}$ and $\lambda_{giou}$ are coefficients of each component.
The training loss is the same as the matching cost except that only performed on matched pairs. The final loss is the sum of all pairs normalized by the number of objects inside the training batch.

\begin{table*}[!t]
\begin{center}
\setlength{\tabcolsep}{2mm}
\begin{tabular}{c|l|c|cccccccc} 
\toprule
Benchmark & 
Method & 
Data & 
MOTA$\uparrow$&
IDF1$\uparrow$& 
MOTP$\uparrow$& 
MT$\uparrow$&  
ML$\downarrow$&  
FP$\downarrow$&
FN$\downarrow$&
IDS$\downarrow$
\\
\midrule
\multirow{14}{*}{MOT17}  

& TubeTK~\cite{pang2020tubetk}  & No  & 63.0 & 58.6 & 78.3 & 31.2 & 19.9 &27060 &177483 & 4137 \\
& ChainedTracker~\cite{peng2020chained}& No & 66.6 &	57.4 & 78.2 & 32.2	& 24.2 &22284 &160491 & 5529 \\
& QuasiDense~\cite{quasidense} & No & 68.7 & 66.3 & 79.0 & 40.6 & 21.9 & 26589	& 146643 & 3378\\

& GSDT~\cite{GSDT} & 5D2 & 73.2 & 66.5 & 80.7 & 41.7 & 17.5 & 26397 & 120666 & 3891\\
& CSTrack~\cite{cstrack} & 5D1 & 74.9 & 72.6 & 80.9 & 41.5 & 17.5 & 23847 & 114303 & 3567\\
& FairMOT~\cite{fairmot} & 5D1 & 73.7 & 72.3 & 81.3 & 43.2 & 17.3 & 27507 & 117477 & 3303\\
& FUFET~\cite{shan2020fgagt} & 5D1 & 76.2 & 68.0 & 81.1 & 51.1 & 13.6 & 32796 & 98475 & 3237 \\
& MLT~\cite{mlt} & 5D1 & 75.3 & 75.5 & 81.7 & 49.3 & 19.5 & 27879 & 109836 & 1719 \\
& CorrTracker~\cite{corrtrack} & 5D1 & 76.5 & 73.6 &81.2 & 47.6 &12.7 &29808 &99510 &3369\\

& CenterTrack~\cite{zhou2020tracking}  & CH & 67.8 & 64.7 & 78.4 & 34.6 & 24.6 & 18489 &160332 & 3039 \\
& TraDeS~\cite{trades} & CH & 69.1 & 63.9 &78.9&36.4 &21.5 & 20892 & 150060	& 3555 \\
& TransMOT~\cite{transmot}  & CH & 76.7 & 75.1 & 82.0 & 51.0 & 16.4 & 36231	& 93150 & 2346  \\
& TransCenter~\cite{transcenter}  & CH & 73.2 & 62.2 & 81.1 & 40.8 & 18.5 & 23112 &  123738 & 4614\\

&\textbf{TransTrack(ours)} & 
CH&
74.5 & 
63.9& 
80.6 & 
46.8 & 
11.3 & 
28323 & 
112137 & 
3663\\

\midrule

\multirow{6}{*}{MOT20} 
& GSDT~\cite{GSDT} & 5D2 & 67.1 & 67.5 & 79.1 & 53.1 & 13.2 & 31507 &	135395 & 3230\\
& CSTrack~\cite{cstrack} & 5D1 & 66.6 & 68.6 & 78.8 & 50.4 & 15.5 & 25404	& 144358 & 3196 \\
& FairMOT~\cite{fairmot} & 5D1 & 61.8 & 67.3 & 78.6 & 68.8 & 7.6 & 103440 &	88901 & 5243\\
& CorrTracker~\cite{corrtrack} & 5D1 & 65.2 & 73.6 & - & 47.6& 12.7& 29808 & 99510 & 3369 \\

& TransCenter~\cite{transcenter}  & CH & 58.3 & 46.8 & 79.7 & 35.7 & 18.6 & 35959 & 174893 & 4947\\
&\textbf{TransTrack(ours)} & 
CH &
64.5 & 
59.2 & 
80.0 & 
49.1 & 
13.6 & 
28566 & 
151377 & 
3565\\
 
\bottomrule

\end{tabular}
\end{center}
\vspace{-5mm}
\caption{\textbf{Evaluation on MOT17 and MOT20 test sets.}
We compare TransTrack with recent methods in private protocol, where external data can be used: CH for CrowdHuman~\cite{shao2018crowdhuman}, 5D1 for the use of 5 extra datasets, including CrowdHuman~\cite{shao2018crowdhuman}, Caltech Pedestrian~\cite{caltechped09,caltechped11}, CityPersons~\cite{citypersons}, CUHK-SYS~\cite{cuhksys}, and PRW~\cite{prw}, 5D2 is the same as 5D1 replacing CroudHuman by ETH~\cite{ethtracking}, NO for using no extra dataset.
}
\label{table5_3}
\end{table*}

\subsection{Inference}
In the inference stage, TransTrack first detects objects on the first frame, where the feature maps are from two copies of the first frame. Then TransTrack operates object propagation and box association for the following frames and finally completes tracklets over the entire video sequence.

We use track rebirth in the inference procedure of TransTrack to enhance robustness to occlusions and short-term disappearing~\cite{bergmann2019tracking,zhou2020tracking,peng2020chained}. Specifically, if a tracking box is unmatched, it keeps as an “inactive” tracking box until it remains unmatched for $K$ consecutive frames. Inactive tracking boxes can be matched to detection boxes and regain their ID.  Following~\cite{zhou2020tracking}, we choose $K = 32$.

\section{Experiments}
To measure the performance of our proposed method,
we conduct experiments on the pedestrian-tracking dataset MOT17~\cite{mot16} and MOT20~\cite{mot20}. In the ablation study, we follow previous practice ~\cite{zhou2020tracking} to split the MOT17 training set into two parts, one for training and the other for validation. We adopt the widely-used MOT metrics set~\cite{bernardin2008evaluating} for quantitative evaluation where multiple object tracking accuracy (MOTA) is the primary metric to measure the overall performance.

\subsection{Implementation details}
We use ResNet-50~\cite{he2016resnet} as the network backbone. The optimizer is AdamW~\cite{loshchilov2017decoupled} and the batch size is set to be 16. The initial learning rate is 2e-4 for the transformer and 2e-5 for the backbone. The weight decay is 1e-4 All transformer weights are initialized with Xavier-init~\cite{glorot2010understanding}, and the backbone model is pretrained on ImageNet~\cite{deng2009imagenet} with frozen batch-norm layers~\cite{ioffe2015batch}. We use data augmentation including random horizontal, random crop, scale augmentation, resizing the input images whose shorter side is by 480 - 800 pixels while the longer side is by at most 1333 pixels. We train the model for 150 epochs and the learning rate drops by a factor of 10 at the 100th epoch. In the ablation study, the model is first pre-trained on CrowdHuman~\cite{shao2018crowdhuman} and then fine-tuned on MOT. When evaluating on the test set, we train our network on combination of CrowdHuman and MOT. More details are discussed in Appendix.

\subsection{MOT17 benchmark}
We evaluate models on MOT17 under the private detector setting. The results We evaluate models on MOT17 under the private detector setting. The results are shown in Table~\ref{table5_3}. TransTrack achieves comparable results with the current state-of-the-art methods, especially in terms of MOTP and FN. The excellent MOTP demonstrates TransTrack can precisely locate objects in the image. The good FN score represents that most objects are successfully detected. Those prove the success of introducing learned object query into the pipeline. As for ID-switch, TransTrack is comparable with the popular trackers, \eg, FairMOT~\cite{fairmot} and CenterTrack~\cite{zhou2020tracking}, which proves the effectiveness of object feature query to associate adjacent frames. Although the ID-switch score of TransTrack is inferior to SOTA methods, it is a promising direction to further improve the overall performance of TransTrack.

\subsection{MOT20 benchmark }
We evaluate models on MOT20 under the private detector setting. The results are shown in Table~\ref{table5_3}. MOT20 includes more crowded scenes than MOT17. Its more severe object occlusion and smaller object size bring more challenges for object detection and tracking. Therefore, all methods show lower performance on MOT20 than on MOT17. But still, TransTrack achieves comparable results with the current state-of-the-art methods on MOT20, in terms of detection metrics and association metrics.

\subsection{Ablation study} 

\subsubsection{Transformer Architecture}
\label{sec:ablation_transformer}
We ablate the effect of Transformer architecture. Four transformer structures are put into comparison. 
\textbf{Transformer} follows the settings of DETR~\cite{DETR} detector, where transformer is built on top of the feature maps of res5 stage~\cite{he2016resnet}. 
\textbf{Transformer-DC5} increases the feature maps resolution. To be precise, we apply dilation convolution to res5 stage and remove a stride from the first convolution of this stage. 
\textbf{Transformer-P3} adopts FPN~\cite{FPN} on the input feature maps. The encoder of the Transformer is directly removed from the whole pipeline for memory limitation. After removing the encoder, the learning rate of the backbone could be raised to the same as transformers. 
Finally, we also tried \textbf{Deformable Transformer}~\cite{deformdetr}, which is a recently proposed architecture to solve the issue of limited resolution in the transformer. Within plausible memory usage, it fuses multiple-scale features into the whole encoder-decoder pipeline and achieves excellent performance in the general object detection dataset.

The quantitative results are shown in Table~\ref{table5_1}. The final performance of \textbf{Transformer} is only 55.4 MOTA. With higher feature resolution, \textbf{Transformer-DC5} yields 3.6 MOTA improvement. However, it also leads to the drawback of dilation convolution, such as big memory usage.  \textbf{Transformer-P3} only outputs close performance as Transformer-DC5, saying that higher resolution than DC5 fails to bring further performance gain. And the reason behind this might be the absence of encoder blocks. At last, \textbf{Deformable Transformer} fuses multiple-scale feature into the whole encoder-decoder pipeline and achieves excellent performance, up to 65.0 MOTA. Therefore, we use Deformable Transformer as the default architecture choice of TransTrack.

\begin{table}[t]
\begin{center}
\setlength{\tabcolsep}{0.5mm}
\begin{tabular}{l c c c c}
\toprule
Architecture & MOTA$\uparrow$ &  FP$\downarrow$ & FN$\downarrow$ & IDs$\downarrow$\\
\midrule
Transformer~\cite{DETR}  & 55.4 & 7.4\% & 35.2\% & 2.0\% \\
Transformer-DC5~\cite{DETR} & 59.0 & 5.2\% & 34.0\% & 1.8\%  \\
Transformer-P3 & 59.3 & 5.1\% & 33.8\% & 1.8\% 	 \\
Deformable Transformer~\cite{deformdetr} & 65.0  & 4.3\%  & 30.3\%  & 0.4\%\\
\bottomrule 
\end{tabular}
\end{center}
\vspace{-5mm}
\caption{\textbf{Ablation study on Transformer architecture.} Original transformer suffers from low feature resolution. Deformable DETR with multi-scale feature input achieves best performance.}
\label{table5_1}
\vspace{-3mm}
\end{table}

\begin{table}[t]
\begin{center}
\setlength{\tabcolsep}{0.5mm}
\begin{tabular}{l c c c c c c c}
\toprule
Query & MOTA$\uparrow$ &  FP$\downarrow$ & FN$\downarrow$ & IDs$\downarrow$\\
\midrule
Obejct query &  58.3 & 4.0\% & 29.7\% & \textcolor{gray}{\textbf{8.0\%}} \\
Track query  & - & 15.6\% & \textcolor{gray}{\textbf{93.8\%}} & 0.3\%  \\
Track query + Object query & 65.0  & 4.3\%  & 30.3\%  & 0.4\%\\
\bottomrule 
\end{tabular}
\end{center}
\vspace{-5mm}
\caption{\textbf{Ablation study on input query.} 
Using only object query obtains limited association performance. Using only track query leads to numerous FN since it misses new-coming objects. By using both object query and track query, the detection and tracking performance are improved. 
}
\label{table5_2}
\end{table}

\begin{figure*}[htbp]
\begin{center}
\includegraphics[width=0.9\linewidth]{ 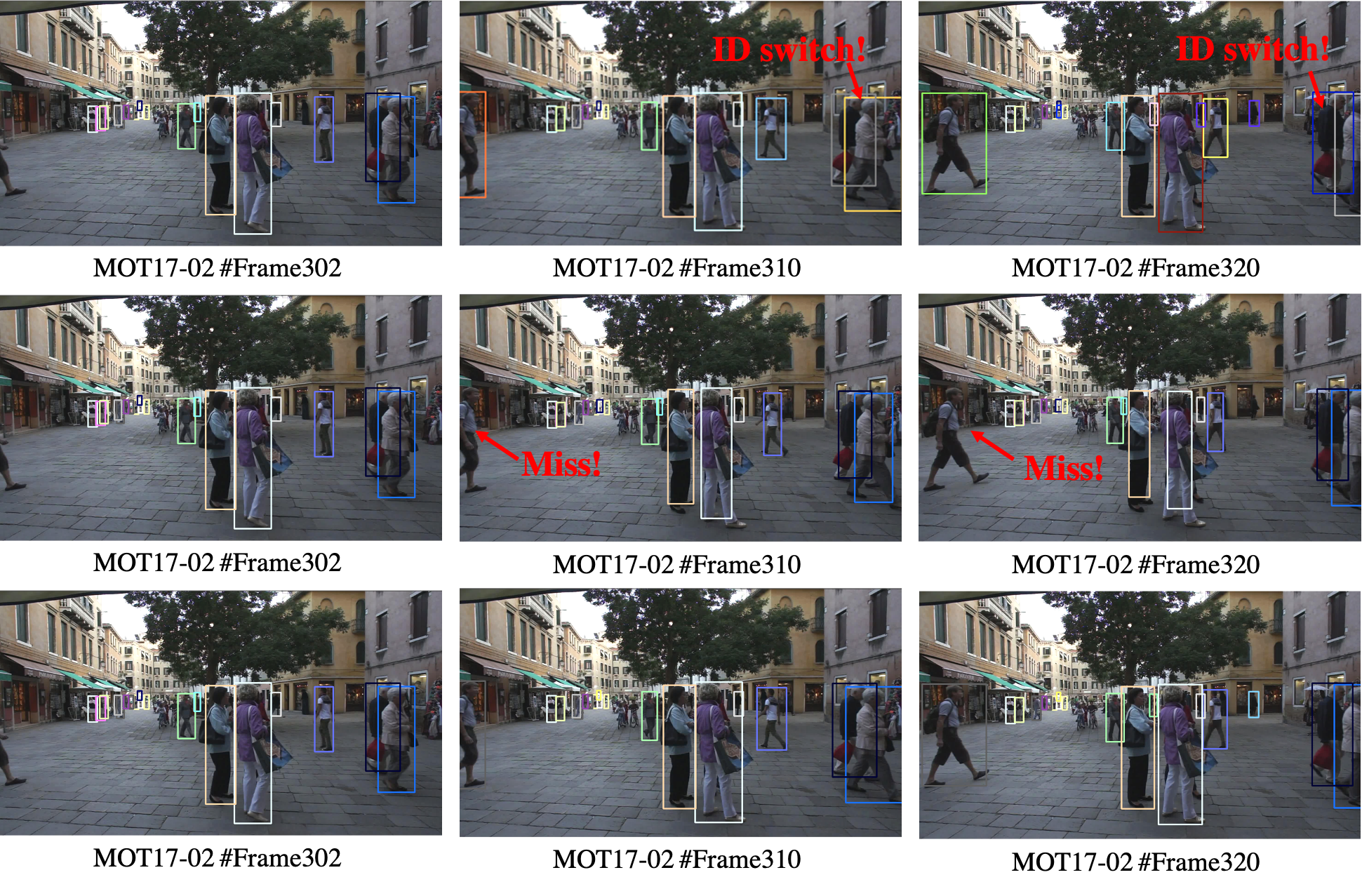}
\caption{\textbf{Visualization of TransTrack with different input query.} 1st row is \textbf{only learned object query}. 2nd row is \textbf{only object feature query from the previous frame}. 3rd row is \textbf{both learned object query and object feature query from the previous frame}. Only learned object query or object feature query from the previous frame causes ID switch case or missing object case. TransTrack takes both as input query and exhibits best detection and tracking performance.} 
\label{fig:vis}
\end{center}
\vspace{-4mm}
\end{figure*}

\subsubsection{Query in Decoder}
\label{sec:ablation_query}
We study the effect of what the input query is used. In the detection task, an input query is generated from the input image only~\cite{DETR, deformdetr}. But in tracking, the knowledge of previously detected objects is expected to be helpful, so we set experiments to compare the model performance when object query and track query are used or absent respectively. The results are reported in Table~\ref{table5_2}.

\noindent\textbf{Only object query.} When only learned object query is input as decoder query, we adopt a naive pipeline where the output detection boxes are associated according to their index in the output set. Surprisingly, this solution achieves a not bad performance by 58.3 MOTA. This is because each object query predicts the object in a certain area on images, and most objects just move around a small distance in the video sequence. However, solely relying on the index in the output set leads to non-negligible wrong matching, especially when the object moves through a long distance. When the object moves around a wide range, this pattern fails easily as visualized in Figure~\ref{fig:vis}.

\noindent\textbf{Only track query.} When only the track query, which is generated from the previous frame, is input to the decoder, we have no common detection results on the image. The visualization in the second row of Figure~\ref{fig:vis} shows that this method is capable to associate the object with a large range of motion. Nevertheless, only the object that appears in the first frame can be tracked successively. For the whole video sequence, most of the objects will be missed and the FN metric collapses as shown in the second row of Table~\ref{table5_2}.

\noindent\textbf{Object query + track query.} As the default setting of TransTrack, both object query and track query are input to the decoder. Now it can handle most failure cases in the previous two cases with the help of the other query. Visualization in Figure~\ref{fig:vis} and performance reported in  Table~\ref{table5_2} prove the giant improvement.

\begin{table}[t]
\centering
\setlength{\tabcolsep}{2.8mm}
\begin{tabular}{l c c c c}
\toprule
Matching  & MOTA $\uparrow$ &  FP$\downarrow$ & FN$\downarrow$ & IDs$\downarrow$\\
\midrule
Previous  & 64.8  & 4.8\%  & 29.8\%  & 0.6\% \\
Current   & 65.0  & 4.3\%  & 30.3\%  & 0.4\% \\
\bottomrule 
\end{tabular}
\caption{\textbf{Ablation study of matching strategy of tracking boxes.} 
\textbf{Previous} indicates directly inheriting the index of track query for box matching on the previous frame.
\textbf{Current} indicates using optimal bipartite matching with current object boxes.
}
\label{table:track_box}
\vspace{-3mm}
\end{table}

\subsubsection{Matching strategy of tracking boxes}
TransTrack builds tracklets based on two sets of detection results and box matching. To emphasize temporal correlation in tracking tasks, it is natural to consider matching tracking boxes with previous frame objects. To ablate the influence of tracking boxes matching, we conduct two strategies. One way is to match initial tracking boxes to previous object boxes by optimal bipartite matching (\textbf{Previous}), in other words, the matching index is directly from the matching index of corresponding track queries. The other strategy is to supervise the output of tracking boxes with optimal bipartite matching to current object boxes (\textbf{Current}). The results are shown in Table~\ref{table:track_box}. The results show that bipartite matching with previous frame objects does not help to void ID switch (0.6\% v.s. 0.4\%). This shows that the inherit the property of the query-key mechanism could well locate the position of the corresponding object on the current frame already.

\begin{table}[t]
\centering
\setlength{\tabcolsep}{2.8mm}
\begin{tabular}{l c c c c}
\toprule
Association  & MOTA $\uparrow$ &  FP$\downarrow$ & FN$\downarrow$ & IDs$\downarrow$\\
\midrule
Hungarian  & 65.0  & 4.3\%  & 30.3\%  & 0.4\% \\
NMS & 65.0  & 4.3\%  & 30.3\%  & 0.4\% \\
\bottomrule 
\end{tabular}
\caption{\textbf{Ablation study of box association.} Two sets of bounding boxes, track boxes and detection boxes, are merged into the desired ordered object set. The results show that Hungarian algorithm and NMS actually have the same effect in this stage.}
\label{table:box_process}
\vspace{-3mm}
\end{table}

\subsubsection{Bounding Box Association}
\label{sec:ablation_post_processing}
 We study the effect of different box association post-processing strategies. We choose the classic Hungarian algorithm~\cite{kuhn1955hungarian} and the NMS merging method used in~\cite{zhou2020tracking,trackformer}. Results are shown in Table~\ref{table:box_process}. It suggests both two strategies show equivalent effect in the box association stage.

\subsection{Comparisons with other trackers}
Two commonly used signals to upgrade a detector to a tracker are motion and appearance features. So ``detector + motion model'' and ``detector + Re-ID'' are widely-used and intuitive methods, thus it is necessary to compare TransTrack with these two models to have a clear idea about how much TransTrack gains from its design except for improvement from the detector it replies on.

\begin{table}[t]
\centering
\setlength{\tabcolsep}{1.0mm}
\begin{tabular}{l c c c c c c }
\toprule
Motion model  & MOTA $\uparrow$ &  FP$\downarrow$ & FN$\downarrow$ & IDs$\downarrow$  & 4xIDs$\downarrow$\\
\midrule
None  &  64.4  & 4.3\%  & 30.3\% &  1.0\% &  1.2\%\\
Kalman filter & 64.9 & 4.3\%  & 30.4\% &  0.4\% & 1.0\%\\
Track query(Ours) & 65.0  & 4.3\%  & 30.3\%  & 0.4\% & 0.5\%\\
\bottomrule 
\end{tabular}
\caption{\textbf{The effect of motion model}. All models use DETR as detectors. For \textbf{None}, object box is associated by IoU similarity. For \textbf{Kalman filter}, the output bounding boxes are processed by Kalman filter.
\textbf{Ours} follows the two-query-set setting where track query is used to associate across-frame objects.
}
\label{table:sort}
\end{table}

\subsubsection{Motion model} 
We combine the widely-used Kalman filter and DETR to build a ``detector + motion model'' tracker. The results are shown in Table~\ref{table:sort}. 
Kalman filter and our method provide similar IDs performance. We explain that the MOT17 dataset is the video sequence of high frame rate (14-30 FPS), where the object motion between two adjacent frames is minor. Therefore, different association methods make no big difference. However, when we sample one frame every 4 frames, the object motion becomes larger, then the improvement brought by the feature query is obvious (0.5\% \vs 1.0\%), shown in the last column of Table~\ref{table:sort}. Similar phenomenons are discussed in~\cite{zhou2020tracking}.

\subsubsection{Re-ID features}
\begin{figure}[!t]
\begin{center}
\includegraphics[width=0.45\textwidth]{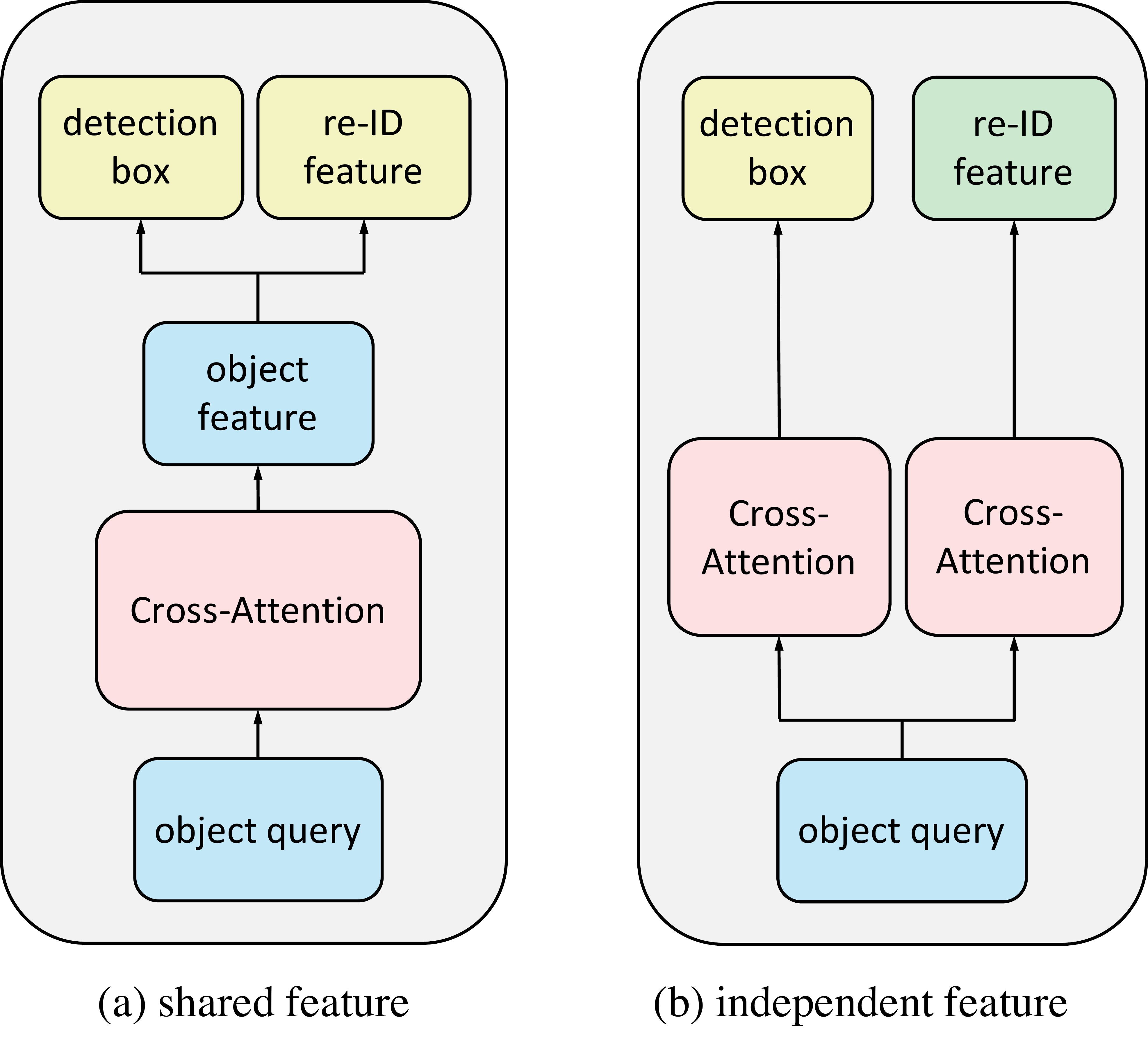}
\vspace{-3mm}
\caption{\textbf{Two designs to introduce Re-ID into DETR.} The left one uses a shared feature from a single cross-attention layer to train detection and re-identification. The right scheme uses two cross-attention layers to generate independent Re-ID features and detection features for the two sources of supervision.
}
\label{fig:reid}
\end{center}
\vspace{-3mm}
\end{figure}

To maintain the joint-detection-and-tracking paradigm, we do not implement an independent Re-ID model but use the Re-ID branch to formulate a ``detector + Re-ID'' tracker. As features generated in the detector have conflicts with appearance-based Re-ID features~\cite{fairmot}, we study the influence of using an independent Re-ID passway, \eg, a cross-attention layer in the decoder. The two patterns are illustrated in Figure~\ref{fig:reid}. The results are included in Table~\ref{table:box_reid}. It agrees that when passed through an independent passway, the Re-ID feature brings better results than using a shared module with detection features. However, the overall MOTA score is not improved against default TransTrack.

\begin{table}[t]
\centering
\setlength{\tabcolsep}{2.0mm}
\begin{tabular}{l c c c c}
\toprule
Re-ID feature   & MOTA $\uparrow$ &  FP$\downarrow$ & FN$\downarrow$ & IDs$\downarrow$\\
\midrule
Shared   &  61.1  & 5.9\%  & 32.3\%  & 0.7\% \\
Independent & 64.7  &3.2\%  & 31.7\%  & 0.4\%\\
None (Ours) & 65.0  & 4.3\%  & 30.3\%  & 0.4\% \\
\bottomrule 
\end{tabular}
\caption{\textbf{The effect of Re-ID features.} When passing Re-ID features to an independent cross-attention layer, the performance is better than using shared cross-attention layer for detection features and Re-ID features. However, this also results in degradation of detector, so the overall performance does not beat default TransTrack.}
\label{table:box_reid}
\vspace{-3mm}
\end{table}

\section{Conclusion}
In this work, we set up a joint-detection-and-tracking MOT pipeline, TransTrack, based on the transformer. It uses the learned object query as input to detects objects and track query, which is the features the from previous frame, to propagate previously detected objects to the following frames. TransTrack is the first work solving MOT in such a paradigm. It achieves a competitive 74.5 MOTA on the MOT17 dataset and 64.5 MOTA on a more challenging MOT20 dataset. We expect it to provide a novel perspective and insight to the MOT community.

\newpage
\appendix
\begin{center}{\bf \Large Appendix}\end{center}\vspace{-2mm}

\section{Training Data}
\label{sec:ablation_pretrain}
We follow the common practice of the state-of-the-art MOT methods~\cite{zhou2020tracking} to train TransTrack on CrowdHuman~\cite{shao2018crowdhuman} first and then fine-tune the model on MOT17. We conduct a comparison to study the effect of the external CrowdHuman data. The result is reported in Table~\ref{table_pre}. Only using the training set of MOT17 merely obtains 61.6 MOTA. When first pre-trained on CrowdHuman then trained on MOT17, the performance achieves 64.8 MOTA. It suggests adding external data boosts the model performance significantly.

\begin{table}[H]
\begin{center}
\setlength{\tabcolsep}{1.1mm}
\begin{tabular}{c c c c c c}
\toprule
Pre-train & 
Fine-tune & 
MOTA$\uparrow$ &  
FP$\downarrow$ & 
FN$\downarrow$ & 
IDs$\downarrow$\\
\midrule
CH & -   & 53.8  & 13.0\%  & 32.3\%  & 1.0\% \\
- & MOT17          & 61.6  & 3.4\%  & 34.2\%  & 0.9\% \\
CH & MOT17 & 65.0  & 4.3\%  & 30.3\%  & 0.4\%\\
\bottomrule 
\end{tabular}
\end{center}
\vspace{-5mm}
\caption{\textbf{Ablation study on pre-training data.} The first row is the model trained only on CrowdHuman dataset. The second row indicates model trained on the training set of MOT dataset only. The third shows the performance when the model is trained on CrowdHuman first and then fine-tuned on MOT dataset. All models are evaluated on the validation set of MOT17 dataset.}
\label{table_pre}
\end{table}

Besides the pre-training data settings, we find the data used for fine-tuning is also critical. We conduct an ablation study on it and the results are shown in Table~\ref{table:table_fine}. Interestingly, fine-tuning on the combination of CrowdHuman and MOT shows better performance than fine-tuning on the MOT dataset only.

\begin{table}[H]
\begin{center}
\setlength{\tabcolsep}{0.5mm}
\begin{tabular}{c | c c c c c c}
\toprule
Dataset &
Pre-train & 
Fine-tune & 
MOTA$\uparrow$ &  
FP$\downarrow$ & 
FN$\downarrow$ & 
IDs$\downarrow$\\
\midrule
\multirow{2}{*}{MOT17}  
& CH & MOT17   & 68.4  & 22137  & 152064  & 3942 \\
& CH & CH+MOT17 & 74.5  & 28323  & 112137  & 3663 \\
\midrule
\multirow{2}{*}{MOT20}  
& CH & MOT20   & 57.4  & 32921  & 184047  & 3705 \\
& CH & CH+MOT20 & 64.5  & 28566  & 151377  & 3565 \\
\bottomrule 
\end{tabular}
\end{center}
\vspace{-5mm}
\caption{\textbf{Ablation study on fine-tuning data.} For each benchmark, the first row is the model fine-tuned only on MOT train dataset. The second row indicates the model fine-tuned on the combination of CrowdHuman and MOT training set. All models are evaluated on the test set of MOT benchmark.}
\label{table:table_fine}
\end{table}

\section{Accuracy vs. Speed}
\label{sec:time_cost}
We analyze the inference speed of TransTrack. The time cost is measured using a single Tesla V100 GPU. Table \ref{table_dec} shows the effect of number of decoders. Increasing decoders hurts inference speed, for example, from 1 decoder to 6, FPS decreases from 15FPS to 10FPS.  However, more decoders significantly boost MOTA performance. Therefore, we choose 6 as the default decoder number. Table \ref{table_size} shows the effect of input image size. Gradually increasing input image size, MOTA performance is saturated when the short-side of the input image is by 800 pixels so we set it as the default setting in TransTrack.

\begin{table}[H]
\begin{center}
\setlength{\tabcolsep}{2.3mm}
\begin{tabular}{c c c c c c}
\toprule
Decoders & 
MOTA$\uparrow$ &  
FP$\downarrow$ & 
FN$\downarrow$ & 
IDs$\downarrow$&
FPS\\
\midrule
1 & 47.0 & 10.0\%  & 40.0\% & 3.0\% & 15\\
3 & 64.3 & 3.3\%   & 31.4\%  & 1.0\% & 12\\
6 & 65.0 & 4.3\%  & 30.3\%  & 0.4\% & 10\\
\bottomrule 
\end{tabular}
\end{center}
\vspace{-5mm}
\caption{\textbf{Ablation study on number of decoders.} Increasing decoders has minor impact on inference time while significantly improves MOTA performance. Therefore, we choose 6 decoders as default.}
\label{table_dec}
\end{table}

\begin{table}[H]
\begin{center}
\setlength{\tabcolsep}{2.0mm}
\setlength{\tabcolsep}{2.3mm}
\begin{tabular}{c c c c c c}
\toprule
Short-side & 
MOTA$\uparrow$ &  
FP$\downarrow$ & 
FN$\downarrow$ & 
IDs$\downarrow$&
FPS\\
\midrule
540 pix  & 62.4 & 3.9\% & 32.8\%  & 0.9\%  & 14\\
800 pix  & 65.0 & 4.3\% & 30.3\%  & 0.4\%  & 10\\
1080 pix & 59.2 & 4.7\% & 35.0\%  & 1.1\%  & 7\\
\bottomrule 
\end{tabular}
\end{center}
\vspace{-5mm}
\caption{\textbf{Ablation study on input image size.} Gradually increasing input image size, MOTA performance is saturated when the short-side of image is 800 pixels. }
\label{table_size}
\end{table}

\newpage
{\small
\bibliographystyle{ieee_fullname}
\bibliography{egbib}
}

\end{document}